\PassOptionsToPackage{unicode}{hyperref}
\PassOptionsToPackage{hyphens}{url}
\PassOptionsToPackage{dvipsnames,svgnames,x11names}{xcolor}
\documentclass[
  11pt,
  letterpaper,
]{article}
\usepackage{xcolor}
\usepackage[margin=1in]{geometry}
\usepackage{amsmath,amssymb}
\usepackage{iftex}
\ifPDFTeX
  \usepackage[T1]{fontenc}
  \usepackage[utf8]{inputenc}
  \usepackage{textcomp} 
\else 
  \usepackage{unicode-math} 
  \defaultfontfeatures{Scale=MatchLowercase}
  \defaultfontfeatures[\rmfamily]{Ligatures=TeX,Scale=1}
\fi
\usepackage{lmodern}
\ifPDFTeX\else
\fi
\IfFileExists{upquote.sty}{\usepackage{upquote}}{}
\IfFileExists{microtype.sty}{
  \usepackage[]{microtype}
  \UseMicrotypeSet[protrusion]{basicmath} 
}{}
\makeatletter
\@ifundefined{KOMAClassName}{
  \IfFileExists{parskip.sty}{%
    \usepackage{parskip}
  }{
    \setlength{\parindent}{0pt}
    \setlength{\parskip}{6pt plus 2pt minus 1pt}}
}{
  \KOMAoptions{parskip=half}}
\makeatother
\usepackage{longtable,booktabs,array}
\usepackage{caption}
\usepackage{calc} 
\usepackage{etoolbox}
\makeatletter
\patchcmd\longtable{\par}{\if@noskipsec\mbox{}\fi\par}{}{}
\makeatother
\usepackage{graphicx}
\makeatletter
\newsavebox\pandoc@box
\newcommand*\pandocbounded[1]{
  \sbox\pandoc@box{#1}%
  \Gscale@div\@tempa{\textheight}{\dimexpr\ht\pandoc@box+\dp\pandoc@box\relax}%
  \Gscale@div\@tempb{\linewidth}{\wd\pandoc@box}%
  \ifdim\@tempb\p@<\@tempa\p@\let\@tempa\@tempb\fi
  \ifdim\@tempa\p@<\p@\scalebox{\@tempa}{\usebox\pandoc@box}%
  \else\usebox{\pandoc@box}%
  \fi%
}
\def\fps@figure{htbp}
\makeatother
\global\tracinglostchars=3
\usepackage{xurl}
\usepackage{seqsplit}
\usepackage{etoolbox}
\usepackage{caption}
\usepackage{float}
\usepackage{needspace}
\AtBeginEnvironment{longtable}{\small}
\usepackage{bookmark}
\IfFileExists{xurl.sty}{\usepackage{xurl}}{} 
\makeatletter
\@ifundefined{xmpquote}{}{}
\makeatother
\hypersetup{
  pdftitle={W-ISA: Transactional Execution of Weight-Space Intervention Programs},
  pdfauthor={Ekrem Aksoy},
  colorlinks=true,
  linkcolor={blue},
  filecolor={Maroon},
  citecolor={Blue},
  urlcolor={blue},
  pdfcreator={LaTeX via pandoc}}

\title{W-ISA: Transactional Execution of Weight-Space Intervention Programs}
\author{Ekrem Aksoy}
\date{November 5, 2026}

\begin{document}
\maketitle
\begin{abstract}
We present W-ISA, a software instruction set and transactional runtime
for executing weight-space intervention programs on pretrained language
models. Shape-checked instructions express parameter updates,
projections, ablations, and rank-one transformations. The runtime
applies these programs within request-scoped execution contexts and
restores saved parameters after completion or interruption.
Activation-based match-action rules connect measured model states to
intervention selection. We evaluate restoration and replay on
SmolLM-135M-Instruct and Qwen2.5-1.5B-Instruct, and compare materialized
rank-one updates with equivalent PyTorch hooks on the latter. In 45
timed requests, both intervention implementations produce identical
greedy sequences at 32, 128, and 512 tokens; their maximum prompt-logit
difference is 0.000184. On the complete WikiText-2 raw test split, the
combined SmolLM attention-head ablation and MLP projection changes
perplexity from 22.2353 to 22.3188, a relative increase of 0.3754\%.
W-ISA separates intervention specification, execution, and measurement,
making parameter-level experiments reproducible and composable.
\end{abstract}


\section{Introduction}\label{introduction}

Interventions on neural networks provide a way to study how internal
computations affect model outputs. Researchers edit factual
associations, add contrastive activation directions, ablate components,
and learn transformations of hidden representations. These techniques
connect a scientific hypothesis to a change in the model's computation.

We explicitly credit Rikiya Takehi, Jayden Teoh, Alex Zhang, and Omar
Khattab for the weight-space programming framing developed in
\emph{Sparks of Weight-Space Programming: LLMs can natively read \&
write neural network weights} {[}10{]}. Their November 2, 2026 article
describes general-purpose LLMs inspecting, constructing, and editing
neural weights through reasoning, including tool-assisted intervention
on a language model. This perspective motivates the execution-layer
question addressed here: how to represent and run weight interventions
with explicit scope, validation, and restoration semantics.

W-ISA addresses the execution of such interventions through an explicit
program representation and a managed parameter-state lifecycle. An
intervention program identifies the weights it touches, specifies the
transformation to apply, and executes within a scope that owns the
corresponding snapshots. The same scope coordinates inference and
restores entry-state weights when it exits.

The central abstraction is \textbf{a weight-space intervention program}.
A request temporarily evaluates a pretrained model in the parameter
state produced by that program. Program execution, activation-based
selection, and evaluation are separate components.

This perspective leads to the following research question:

\begin{quote}
How can explicit weight-space programs execute within a shared
language-model instance while preserving request-scoped state, exact
parameter restoration, and reproducible measurement?
\end{quote}

The implementation brings together a shape-checked instruction
representation, direct parameter mutation, nested snapshot scopes, and
activation-based match-action selection. Our contribution is this
execution abstraction and its concrete operational semantics,
accompanied by a reproducible measurement workflow.

\section{Related Work}\label{related-work}

\subsection{Weight-space programming and circuit
structure}\label{weight-space-programming-and-circuit-structure}

Takehi et al.~{[}10{]} investigate LLM-native reading and writing of
network weights and provide detailed experiment logs. We adopt their
weight-space programming perspective as a conceptual foundation. W-ISA's
contribution is a structured instruction and transactional execution
layer for supplied interventions.

Elhage et al.~{[}11{]} analyze the linear structure of the residual
stream and the QK/OV decomposition of attention heads. Their treatment
of head outputs as additive contributions provides background for
addressing individual output-projection slices. This work is also cited
in Takehi et al.'s discussion of weight-based circuit interpretation.

Tracr {[}13{]}, linked in Takehi et al.'s introductory discussion,
compiles human-readable programs into decoder-only transformers with
known structure. It provides a programming-to-weights reference point:
Tracr constructs transformer models from programs, while W-ISA compiles
intervention instructions bound to the parameters of an existing model.

\subsection{Weight editing}\label{weight-editing}

ROME uses rank-one model editing to update factual associations in
transformer feed-forward weights {[}1{]}. MEMIT extends direct memory
editing to collections of factual associations {[}2{]}. These methods
motivate a distinction between the algorithm that constructs an edit and
the runtime that applies it. W-ISA exposes supplied updates and
algebraic operations through a common instruction interface.

\subsection{Activation interventions}\label{activation-interventions}

Activation Addition derives steering directions from contrasting prompt
activations and applies them during inference {[}3{]}. Contrastive
Activation Addition averages differences between positive and negative
examples and adds the resulting directions to hidden representations
{[}4{]}. ReFT learns task-specific transformations of representations in
a frozen base model {[}5{]}. W-ISA's profiler similarly exposes
activation measurements, while its instruction runtime applies explicit
transformations to parameter tensors.

Arditi et al.~{[}12{]} study a residual-stream direction mediating
refusal behavior and examine interventions on that direction. We cite
this work as a concrete precedent for directional behavioral
interventions. The projections reported here use measured input-space
directions at the selected MLP down-projection.

\subsection{Intervention software}\label{intervention-software}

pyvene provides declarative configurations, static and trainable
interventions, and sequential or parallel intervention schemes {[}6{]}.
NNsight provides temporary tracing interventions, persistent edit
definitions, and activation-dependent conditional execution {[}7, 8{]}.
These systems are direct comparison points for intervention
representation, composition, and lifecycle management.

W-ISA focuses on parameter-level programs with owned snapshots,
finite-value and shape checks, source-order execution, and
request-scoped restoration. Its research framing is the combination of
these mechanisms into a weight-state execution contract. Section 8
reports a measured comparison with baseline PyTorch inference and an
equivalent forward hook.

\section{Weight-Space Programs}\label{weight-space-programs}

Let \(f_\theta\) denote a pretrained model with parameter state
\(\theta\). A program

\[P=(I_1,I_2,\ldots,I_k)\]

is an ordered sequence of instructions. Each instruction names a
parameter, optionally selects a rectangular tensor region, and provides
an opcode with scalar or tensor operands. Compilation resolves parameter
names and checks the selected dimensions against operand shapes.

For a selected matrix \(W\in\mathbb{R}^{m\times n}\), the instruction
set is:

\Needspace{15.0\baselineskip}

{\def\LTcaptype{none} 
\begin{longtable}[]{@{}
  >{\raggedright\arraybackslash}p{(\linewidth - 2\tabcolsep) * \real{0.5000}}
  >{\raggedright\arraybackslash}p{(\linewidth - 2\tabcolsep) * \real{0.5000}}@{}}
\toprule\noalign{}
\begin{minipage}[b]{\linewidth}\raggedright
Instruction
\end{minipage} & \begin{minipage}[b]{\linewidth}\raggedright
Operation
\end{minipage} \\
\midrule\noalign{}
\endhead
\bottomrule\noalign{}
\endlastfoot
\texttt{ADD} & \(W\leftarrow W+\alpha D\) \\
\texttt{SCALE} & \(W\leftarrow aW\) \\
\texttt{RANK1} & \(W\leftarrow W+\alpha uv^T\) \\
\texttt{PROJECT\_OUT} & \(W\leftarrow W-(W\hat v)\hat v^T\), where
\(\hat v=v/\lVert v\rVert_2\) \\
\texttt{WRITE\_KEY} & \(W\leftarrow W+(y-Wk)k^T/(k^Tk)\) \\
\texttt{PRUNE} & \(W_{ij}\leftarrow0\) where \(|W_{ij}|<t\) \\
\end{longtable}
}

\texttt{ADD}, \texttt{SCALE}, and \texttt{PRUNE} also operate on other
dense parameter shapes. Projection and key-writing instructions use
nonzero input directions. Named operands are detached and copied into
CPU FP64 storage during compilation.

The execution transform includes storage rounding after every
instruction. If \(T_i\) is the mathematical operation and \(Q_i\)
converts its result to the target parameter dtype, the effective
sequence is

\[T_P=Q_k\circ T_k\circ\cdots\circ Q_1\circ T_1.\]

This makes instruction order and intermediate precision part of the
execution semantics. Reports record the norm and number of changed
elements after each storage conversion.

\subsection{A connection to activation-space
execution}\label{a-connection-to-activation-space-execution}

An input-space projection can be expressed either as an activation
transformation or a weight update:

\[W' = W\left(I - \hat{v}\hat{v}^{T}\right)
\quad\Longrightarrow\quad
W'h = W\left(h - \hat{v}\left(\hat{v}^{T}h\right)\right).\]

Likewise, a rank-one update satisfies

\[(W+uv^T)h=Wh+u(v^Th).\]

These identities connect weight-space instructions to equivalent linear
interventions on a layer's computation in exact arithmetic. They provide
a basis for comparing activation-side execution with materializing a
weight update for the duration of a request. Such comparisons can
measure arithmetic differences after dtype rounding alongside
application cost and per-token overhead.

\section{Transactional Parameter
State}\label{transactional-parameter-state}

\subsection{Scoped execution}\label{scoped-execution}

For a request \(x\) and program \(P\), the intended execution contract
is

\[\operatorname{ExecuteScope}(\theta,x,P)
=\left(f_{T_P(\theta)}(x),\theta\right).\]

The first component is the request result produced with the intervention
active. The second component is the restored parameter state at scope
exit. This contract concerns runtime-managed weights, with parameter
bindings and storage layouts fixed during the scope.

Before a write, the runtime captures the affected parameter in its
original storage dtype. Instruction arithmetic runs on CPU in FP64.
Results are checked for finite values, converted to the original dtype,
and copied into the existing parameter object. The model remains on its
selected device.

\texttt{temporary()} holds a reentrant lock shared by runtime instances
wrapping the same model object. It registers touched parameters in
active scopes and restores their saved values through Python's
\texttt{finally} mechanism. Completion, Python exceptions, and
cancellation follow the same scope-exit path. A separate
\texttt{execute()} transaction also restores completed writes when a
later instruction fails.

\subsection{Operational reversibility}\label{operational-reversibility}

Some instructions are information-losing transformations: zero scaling
removes a selected component, and projection removes a direction. W-ISA
achieves \textbf{operational reversibility} by retaining the entry-state
parameter values.

For a scope with saved parameter set \(S\), the restoration condition is

\[\forall p\in S:\quad
\operatorname{bytes}(p_{\mathrm{exit}})
=\operatorname{bytes}(p_{\mathrm{entry}}).\]

Restoration follows from copying the original-dtype snapshots back to
their bound parameters. It uses saved state rather than a numerical
inverse of the intervention. Nested scopes record each parameter's first
pre-write value within that scope; an inner exit restores the inner
entry state, followed by the outer entry state when the outer scope
exits.

\subsection{Isolation and device
execution}\label{isolation-and-device-execution}

The shared model lock serializes participating requests for the lifetime
of the intervention. Matched and unmatched requests use the same lock.
Generation begins with a fresh KV cache after intervention selection.

The implementation accepts CPU, MPS, and indexed CUDA parameter devices.
Edit arithmetic remains on CPU in FP64. Touched accelerator devices are
synchronized before snapshots and after writes, and restoration waits
for pending device work. This separates the model's execution device
from the edit arithmetic.

The software test run recorded on November 4, 2026 passed 66 tests in
the CPU/MPS environment. These cover instruction arithmetic, dtype
conversion, parameter slices, restoration, cancellation, request
isolation, model integration and the hook comparison runner.
Device-selection and synchronization-routing tests use mocked backend
discovery. The pretrained measurements in Sections 6--8 use MPS.

\subsection{Snapshot and workspace
scaling}\label{snapshot-and-workspace-scaling}

For touched parameter tensors \(p_j\) with element sizes \(s_j\), one
scope's CPU snapshot storage is

\[M_{\mathrm{snapshot}} = \sum_{j} s_j\,\lvert p_j\rvert.\]

Snapshots cover full target parameters, including when an instruction
selects a smaller slice. The selected region determines arithmetic
workspace. If its largest element count is \(M\) and the compiled
operands contain \(A\) elements, the FP64 working storage is
\(O(8M+8A)\) with several dense temporaries during evaluation. A
preceding undo point and nested scopes retain snapshots of touched
parameters; scopes can share a backup allocation when their first-write
entry state coincides. Total model parameter storage is a separate term.

For a 100-million-element target tensor, a single FP16 snapshot occupies
200 MB and one FP64 working buffer occupies 800 MB. These are byte-count
calculations for the stated tensor and dtype. On the measured Qwen
target, the 1536×8960 down-projection contains 13,762,560 elements: its
FP32 snapshot is 52.5 MiB, one FP64 buffer is 105 MiB, and the rank-one
operand pair occupies 82 KiB.

\section{Activation-Based
Match-Action}\label{activation-based-match-action}

A profiler captures an input or output activation at a named module,
using last-token or mean-token pooling. Given positive and negative
prompt sets, it computes

\[v_{\mathrm{diff}}=\mu_+-\mu_-.\]

A rule associates a capture site, direction, threshold, and intervention
program. For request \(x\), a score is obtained from the activation
measured in the request-entry parameter state:

\[s(x)=\frac{h_\theta(x)^Tv}{\lVert h_\theta(x)\rVert_2\lVert v\rVert_2}.\]

The first rule in caller-supplied order satisfying \(s(x)>\tau\) selects
the program. The controller measures rule conditions before applying a
patch. Rules at the same capture site reuse the captured activation.

The resulting lifecycle is: capture, select, snapshot, apply, generate,
and restore. Threshold calibration uses a designated calibration set.
The profile stores its capture location, pooling method, model revision,
sample counts, and direction checksum.

\section{Experimental Setup}\label{experimental-setup}

We use \texttt{\seqsplit{HuggingFaceTB/SmolLM-135M-Instruct}}, revision
\texttt{\seqsplit{b5f4b394abe60fe0e4d666139e57bfce520302a9}}, with
134,515,008 parameters, and
\texttt{\seqsplit{Qwen/Qwen2.5-1.5B-Instruct}}, revision
\texttt{\seqsplit{989aa7980e4cf806f80c7fef2b1adb7bc71aa306}}, with
1,543,714,304 parameters. Recorded accelerator experiments use FP32
weights on Apple MPS and CPU FP64 edits. The environment is ARM64 macOS
with 24 GiB unified memory, Python 3.12.9, PyTorch 2.15.0, Transformers
5.22.1, and NumPy 2.5.4. The corpus and conditional-rule measurements
use SmolLM; the larger-model roundtrip and execution comparison use
Qwen.

\subsection{Head ablation and
projection}\label{head-ablation-and-projection}

The tenth SmolLM transformer block, indexed as \texttt{layers{[}9{]}},
has nine query heads, three KV heads, and 64 dimensions per query head.
\texttt{SCALE} with factor zero targets columns \texttt{{[}0,64)} of its
576×576 attention output projection. This removes query head 0's output
contribution while preserving the other head columns.

\texttt{PROJECT\_OUT} targets the same block's 576×1536 MLP
down-projection. Its direction is the measured down-projection input at
the last token of the first evaluation prompt, captured before editing.
The direction is held fixed across the experiment.

Four fixed prompts are evaluated through baseline, baseline repeat,
edited, edited repeat, restored, and replayed phases. Each generation
uses greedy decoding with up to 48 new tokens and a fresh cache. Reports
retain generated token IDs, text, next-token logits, parameter digests,
and arithmetic traces.

\subsection{Corpus evaluation}\label{corpus-evaluation}

The complete \texttt{Salesforce/wikitext}, \texttt{wikitext-2-raw-v1},
test split contains 4,358 rows {[}9{]}. Rows are joined with two
newlines and tokenized as raw text. The resulting 304,986-token sequence
supplies 304,985 prediction targets, each scored once.

Evaluation uses 512 input-token windows and stride 256. The 1,191
windows retain per-window NLL sums and token counts. FP32 token losses
are accumulated in CPU FP64, and perplexity is

\[\operatorname{PPL}=\exp\left(\frac{\sum_{t=1}^{N}-\log p_\theta(x_t\mid x_{<t})}{N}\right),\]

where the conditioning context is restricted to the documented rolling
window and \(N=304{,}985\).

\subsection{Contrastive profile and conditional
requests}\label{contrastive-profile-and-conditional-requests}

A handwritten formal/casual fixture uses five topic pairs for profiling,
five different topic pairs for threshold calibration, and six further
topic pairs for evaluation. Each topic receives matched formal and
casual instruction prefixes. The profiler captures the last prompt token
at the tenth block's MLP down-projection input.

Calibration selects the cosine threshold that maximizes balanced
accuracy, with tie-breaking toward lower false-positive rate and then
the larger threshold. The held-out evaluation contains twelve prompts.

\section{Results}\label{results}

\subsection{Restoration and
repeatability}\label{restoration-and-repeatability}

Head ablation changed 36,864 entries. The MLP projection changed 884,666
of 884,736 entries after FP32 rounding. Other head columns and
target-external parameter digests were preserved.

For the normalized captured direction, the output norm changed from
6.4653542059 to \(1.4276433\times10^{-7}\). The FP32 stored projection
matched an independent NumPy implementation with maximum absolute
difference 0.0 in the recorded run.

All four prompts changed their next-token logits, and three changed
their generated token sequences. Undo restored the original weights and
recorded outputs; replay restored the edited state. Two separately
launched processes matched in saved logits, token sequences, direction
values, and parameter digests. The four recorded combined-edit
transactions took 9.2--14.6 ms, excluding inference.

\subsection{Four-condition perplexity}\label{four-condition-perplexity}

\Needspace{12.0\baselineskip}

{\def\LTcaptype{none} 
\begin{longtable}[]{@{}lrr@{}}
\toprule\noalign{}
Condition & Perplexity & Relative change \\
\midrule\noalign{}
\endhead
\bottomrule\noalign{}
\endlastfoot
Original & 22.23531898 & --- \\
Head ablation & 22.31904706 & +0.37655\% \\
MLP projection & 22.23131240 & -0.01802\% \\
Combined & 22.31878257 & +0.37536\% \\
\end{longtable}
}

Each condition uses identical tokens and evaluation settings. Target
parameters were restored between conditions. A full pass took
approximately 61 seconds. Independent reconstruction from saved window
losses reproduced the aggregate perplexities and target counts.

\subsection{Contrastive direction and conditional
execution}\label{contrastive-direction-and-conditional-execution}

The formal-minus-casual direction has norm 0.2804317523. Calibration
selected

\[{\tau} = -0.018948859620056394.\]

\Needspace{12.0\baselineskip}

{\def\LTcaptype{none} 
\begin{longtable}[]{@{}lrr@{}}
\toprule\noalign{}
Trigger result & Calibration & Held-out evaluation \\
\midrule\noalign{}
\endhead
\bottomrule\noalign{}
\endlastfoot
True positives & 5/5 & 4/6 \\
False positives & 0/5 & 0/6 \\
False negatives & 0 & 2 \\
Balanced accuracy & 1.0 & 0.83333 \\
\end{longtable}
}

Unconditional projection of this direction changed three of twelve
generated responses, all with casual-style instructions. The mean
formal-completion preference margin changed from 0.34645676 to
0.34226227. The difference between formal-instruction and
casual-instruction margins changed from 0.21085805 to 0.22222082. Full
WikiText-2 perplexity changed from 22.23531898 to 22.23619810, or
+0.0039537\%.

Conditional execution selected the patch for four prompts and retained
original weights for eight. Maximum absolute next-token logit changes on
the matched prompts were 0.265463, 0.307559, 0.313648, and 0.232894. All
twelve conditional requests retained their original greedy token
sequences.

Repeated requests, injected Python errors, cancellation, early closure
of a synthetic stream, and two threaded requests sharing the same model
completed with the recorded restoration checks satisfied. The threaded
execution used the shared model lock to serialize access. Before/after
full-model parameter digests matched.

\Needspace{12.0\baselineskip}

{\def\LTcaptype{none} 
\begin{longtable}[]{@{}lr@{}}
\toprule\noalign{}
Request timing component & Mean \\
\midrule\noalign{}
\endhead
\bottomrule\noalign{}
\endlastfoot
Activation profiling & 13.288 ms \\
Apply phase, matched requests & 10.635 ms \\
Control overhead across twelve requests & 17.493 ms \\
Entire request including inference callback & 442.535 ms \\
\end{longtable}
}

Control overhead is total request time minus inference callback time. It
includes profiling, application, restoration, and lock wait. The
callback computes a diagnostic next-token forward and generates up to 48
tokens.

\section{Larger-Model Validation and Execution
Comparison}\label{larger-model-validation-and-execution-comparison}

\subsection{Qwen2.5-1.5B roundtrip}\label{qwen2.5-1.5b-roundtrip}

The Qwen model has 28 blocks, hidden size 1536, 12 query heads, two KV
heads, and a 128-dimensional head output. We repeat the head-ablation
and MLP-projection program at zero-based layer 9. Head 0 occupies
output-projection columns \texttt{{[}0,128)}. The MLP down-projection
has shape \(1536 \times 8960\). The two full-target snapshots together
occupy 61.5 MiB in FP32.

The head edit changed 196,608 entries and the projection changed
13,753,919 entries after FP32 rounding. The direction-response norm
changed from 1.21802443 to \(3.07455\times10^{-8}\), with relative
residual \(2.52421\times10^{-8}\). The NumPy reference and stored
projection matched with maximum absolute difference 0.0. Initial
application took 130.425 ms and replay took 88.601 ms in this run.

All four prompts changed their next-token logits; three changed
generated text. The arithmetic prompt returned \texttt{42} in both
states. Original/edited repeats, undo, replay, parameter identities and
full-model weight digests satisfied their equality checks. The
experiment records retain the corresponding logits, token sequences and
parameter digests.

\subsection{Equivalent rank-one
execution}\label{equivalent-rank-one-execution}

We compare original inference, W-ISA materialization, and a PyTorch
forward hook implementing \(Wh+\alpha u(v^Th)\), on the Qwen layer-9 MLP
down-projection. The CPU FP64 vectors are normalized elementwise sine
and cosine sequences; \(\alpha=0.1\). The W-ISA path materializes
\(W+\alpha uv^T\) in FP32. The hook retains the base weights and
performs its vector operations in the model's FP32 dtype on MPS.

One fixed matrix-multiplication tutorial prompt encodes to 51 tokens.
The decoder produces exactly 32, 128, or 512 greedy tokens, continuing
the fixed-length workload through any EOS token. Each request begins
with a fresh cache and synchronizes at each generated token. Each
mode/length receives one warm-up and five timed repetitions; execution
order rotates between repetitions. This yields 45 timed requests and 9
warm-up requests.

Setup/application and restoration are included in total request time.
TTFT is measured from request-scope entry to the first synchronized
token. Decode time per token covers the remaining \(L-1\) steps. Model
loading, program compilation, pre-benchmark equivalence validation and
post-request checksum calculation occur outside timed requests. Runtime
target checks, finite-value checks and edit-report calculations remain
inside the application measurement.

The materialized weights match an independent NumPy reference exactly
after FP32 conversion. Across prompt positions, hook versus materialized
logits have maximum absolute difference 0.000183761 and RMS difference
0.000016225, with argmax agreement 1.0. The materialized intervention
changes prompt logits from baseline by up to 0.0411067. All timed token
sequences repeat exactly within each mode, and all three modes produce
the same token sequence at every measured length. Each request restores
the target weight checksum.

\Needspace{19.5\baselineskip}

{\def\LTcaptype{none} 
\begin{longtable}[]{@{}
  >{\raggedright\arraybackslash}p{(\linewidth - 8\tabcolsep) * \real{0.1667}}
  >{\raggedright\arraybackslash}p{(\linewidth - 8\tabcolsep) * \real{0.1667}}
  >{\raggedleft\arraybackslash}p{(\linewidth - 8\tabcolsep) * \real{0.2222}}
  >{\raggedleft\arraybackslash}p{(\linewidth - 8\tabcolsep) * \real{0.2222}}
  >{\raggedleft\arraybackslash}p{(\linewidth - 8\tabcolsep) * \real{0.2222}}@{}}
\toprule\noalign{}
\begin{minipage}[b]{\linewidth}\raggedright
Generated tokens
\end{minipage} & \begin{minipage}[b]{\linewidth}\raggedright
Mode
\end{minipage} & \begin{minipage}[b]{\linewidth}\raggedleft
Median request (s)
\end{minipage} & \begin{minipage}[b]{\linewidth}\raggedleft
Median TTFT (ms)
\end{minipage} & \begin{minipage}[b]{\linewidth}\raggedleft
Median decode (ms/token)
\end{minipage} \\
\midrule\noalign{}
\endhead
\bottomrule\noalign{}
\endlastfoot
32 & Original & 1.1858 & 50.644 & 36.579 \\
32 & W-ISA & 1.2312 & 99.210 & 36.426 \\
32 & Hook & 1.1868 & 50.776 & 36.626 \\
128 & Original & 4.7445 & 50.576 & 36.941 \\
128 & W-ISA & 4.8527 & 108.252 & 36.894 \\
128 & Hook & 4.7873 & 52.161 & 37.268 \\
512 & Original & 18.7117 & 50.941 & 36.515 \\
512 & W-ISA & 19.3735 & 166.428 & 37.579 \\
512 & Hook & 19.4961 & 51.137 & 38.049 \\
\end{longtable}
}

\begin{figure}
\centering
\pandocbounded{\includegraphics[keepaspectratio,alt={Qwen2.5-1.5B
execution comparison. Points are medians of five requests; error bars
show the interquartile range. Left: complete request duration including
setup and restoration. Right: synchronized decode duration per generated
token after the first.}]{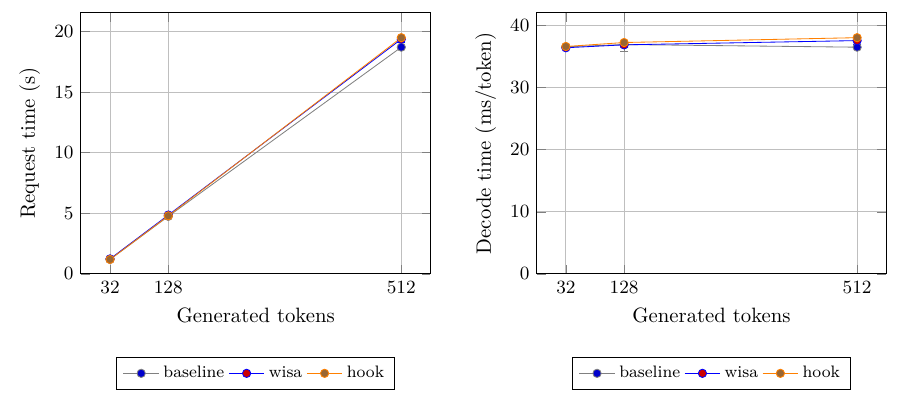}}
\caption{Qwen2.5-1.5B execution comparison. Points are medians of five
requests; error bars show the interquartile range. Left: complete
request duration including setup and restoration. Right: synchronized
decode duration per generated token after the first.}
\end{figure}

The measured W-ISA application medians are 48.866, 56.059 and 115.142 ms
for the three length groups; restoration medians are 2.500, 2.413 and
2.624 ms. Hook setup medians are 0.737, 0.960 and 0.922 ms. The
application work targets the same matrix in each group; the per-request
measurements retain the allocation state and timing variation of the
rotating run.

\subsection{Memory measurements}\label{memory-measurements}

Process RSS and MPS active/driver allocation counters were sampled every
10 ms through each timed request. The report records starting,
sampled-peak and ending values. RSS and MPS counters describe
overlapping aspects of unified-memory accounting and are reported
separately.

\Needspace{10.5\baselineskip}

{\def\LTcaptype{none} 
\begin{longtable}[]{@{}
  >{\raggedright\arraybackslash}p{(\linewidth - 6\tabcolsep) * \real{0.2000}}
  >{\raggedleft\arraybackslash}p{(\linewidth - 6\tabcolsep) * \real{0.2667}}
  >{\raggedleft\arraybackslash}p{(\linewidth - 6\tabcolsep) * \real{0.2667}}
  >{\raggedleft\arraybackslash}p{(\linewidth - 6\tabcolsep) * \real{0.2667}}@{}}
\toprule\noalign{}
\begin{minipage}[b]{\linewidth}\raggedright
Mode
\end{minipage} & \begin{minipage}[b]{\linewidth}\raggedleft
Maximum sampled RSS (MiB)
\end{minipage} & \begin{minipage}[b]{\linewidth}\raggedleft
Maximum MPS active allocation (MiB)
\end{minipage} & \begin{minipage}[b]{\linewidth}\raggedleft
Maximum RSS increase over request start (MiB)
\end{minipage} \\
\midrule\noalign{}
\endhead
\bottomrule\noalign{}
\endlastfoot
Original & 1446.53 & 5975.53 & 1.67 \\
W-ISA & 1446.84 & 5975.52 & 498.94 \\
Hook & 1445.91 & 5975.60 & 1.64 \\
\end{longtable}
}

The model's parameter tensors occupy 6,174,857,216 bytes. The target
snapshot and FP64 buffer sizes in Section 4.4 follow directly from
tensor dimensions. The observed RSS increments also include the
allocator and dense intermediate buffers used during materialization.
JSON records retain every sampled request maximum and the MPS driver
counters.

\subsection{Execution alternatives and companion
controller}\label{execution-alternatives-and-companion-controller}

The materialized path pays snapshot, update and restoration costs around
standard dense inference. The hook path pays setup cost and evaluates
\(\alpha u(v^Th)\) during each layer invocation. These measured forms
separate one-time mutation cost from repeated intervention arithmetic.

For a serving extension, the hook formulation gives a request-local
representation of \(u\), \(v\) and \(\alpha\) alongside shared base
weights. A backend can route those factors by request identity using
dynamic adapters. Such routing operates on separate intervention data; a
tensor view of the base matrix shares its underlying storage.

The companion Neural Control Plane (NCP) manuscript {[}14{]} addresses
proposal generation, telemetry, verification and admission of
intervention policies. W-ISA provides the scoped execution contract used
by that controller; NCP provides the policy-selection loop. The
manuscripts share the runtime and identify their experiments separately.

\section{Reproducibility}\label{reproducibility}

The implementation comprises the instruction compiler, runtime,
profiler, controller, offline tests, and experiment runners. Model
checkpoints are obtained by model identifier or local path; the WikiText
downloader records the corpus checksum and row count. The saved source
programs and numeric operand archives identify the actual interventions
used in the measurements.

The roundtrip records retain original and fresh-process outputs, raw
logits, operands, and the intervention program. Corpus and
conditional-request records retain per-window losses, activation
profiles and routing decisions. The test suite checks independent matrix
calculations, instruction ordering, dtype conversion, snapshot
restoration, request isolation, and device routing.

Sections 6 and 8 specify the model revisions, devices, dtypes,
intervention locations and decoding protocols used in the experiments.
The benchmark records each request's timing and memory measurements
together with warm-up and repetition settings, and exports CSV
measurements and the figure's SVG, PGFPlots source and PDF.

Code is distributed under the MIT License, copyright 2026 Ekrem Aksoy.
Model and dataset materials retain their upstream licenses and
attribution.

\section{Conclusion}\label{conclusion}

W-ISA treats model parameters as managed execution state. An explicit
intervention program defines the transformation; a request scope
controls its lifetime; snapshots restore its entry state; and a
measurement layer records its effects and cost. The resulting
abstraction connects weight editing, mechanistic intervention, and
runtime state management within a small executable system.

\section*{References}\label{references}
\addcontentsline{toc}{section}{References}

\small\raggedright\setlength{\parskip}{3pt}

{[}1{]} Kevin Meng, David Bau, Alex Andonian, and Yonatan Belinkov.
\emph{Locating and Editing Factual Associations in GPT}. NeurIPS, 2022.
\url{https://arxiv.org/abs/2202.05262}

{[}2{]} Kevin Meng, Arnab Sen Sharma, Alex Andonian, Yonatan Belinkov,
and David Bau. \emph{Mass-Editing Memory in a Transformer}.
arXiv:2210.07229. \url{https://arxiv.org/abs/2210.07229}

{[}3{]} Alexander Matt Turner et al.~\emph{Steering Language Models With
Activation Engineering}. arXiv:2308.10248.
\url{https://arxiv.org/abs/2308.10248}

{[}4{]} Nina Panickssery et al.~\emph{Steering Llama 2 via Contrastive
Activation Addition}. arXiv:2312.06681.
\url{https://arxiv.org/abs/2312.06681}

{[}5{]} Zhengxuan Wu et al.~\emph{ReFT: Representation Finetuning for
Language Models}. arXiv:2404.03592.
\url{https://arxiv.org/abs/2404.03592}

{[}6{]} Zhengxuan Wu et al.~\emph{pyvene: A Library for Understanding
and Improving PyTorch Models via Interventions}. arXiv:2403.07809.
\url{https://arxiv.org/abs/2403.07809}

{[}7{]} NNsight documentation. \emph{Model Editing}. Accessed November
3, 2026. \url{https://nnsight.net/features/7_model_editing/}

{[}8{]} NNsight documentation. \emph{Conditional Interventions}.
Accessed November 3, 2026. \url{https://nnsight.net/vllm/conditional/}

{[}9{]} Stephen Merity, Caiming Xiong, James Bradbury, and Richard
Socher. \emph{Pointer Sentinel Mixture Models}. arXiv:1609.07843.
\url{https://arxiv.org/abs/1609.07843}. Dataset distribution:
\url{https://huggingface.co/datasets/Salesforce/wikitext}

{[}10{]} Rikiya Takehi, Jayden Teoh, Alex Zhang, and Omar Khattab.
\emph{Sparks of Weight-Space Programming: LLMs can natively read \&
write neural network weights}. Blog post, November 2, 2026.
\url{https://rikiyat.github.io/blog/weight-space-programming/}

{[}11{]} Nelson Elhage et al.~\emph{A Mathematical Framework for
Transformer Circuits}. Transformer Circuits Thread, December 22, 2021.
\url{https://transformer-circuits.pub/2021/framework/index.html}

{[}12{]} Andy Arditi, Oscar Obeso, Aaquib Syed, Daniel Paleka, Nina
Panickssery, Wes Gurnee, and Neel Nanda. \emph{Refusal in Language
Models Is Mediated by a Single Direction}. arXiv:2406.11717, 2024.
\url{https://arxiv.org/abs/2406.11717}

{[}13{]} David Lindner, János Kramár, Sebastian Farquhar, Matthew Rahtz,
Thomas McGrath, and Vladimir Mikulik. \emph{Tracr: Compiled Transformers
as a Laboratory for Interpretability}. NeurIPS, 2023.
\url{https://arxiv.org/abs/2301.05062}

{[}14{]} Ekrem Aksoy. \emph{NCP: Verifier-Guided Optimization of
Weight-Space Control Policies}. Companion manuscript, 2026.

\end{document}